\documentclass[preprint,12pt]{elsarticle}

\usepackage{amssymb}
\usepackage[utf8]{inputenc}
\usepackage[T1]{fontenc}
\usepackage{url,lineno,microtype}
\usepackage[onehalfspacing]{setspace}
\usepackage{booktabs}
\usepackage{hyperref}
\usepackage{multirow}
\usepackage{threeparttable}
\usepackage{color}
\usepackage{times}
\usepackage{epsfig}
\usepackage{graphicx}
\usepackage{amsmath}
\usepackage{amsthm}
\usepackage{amssymb}
\usepackage{algorithm}
\usepackage{algorithmic}
\usepackage{makecell}
\usepackage{subfigure}

\journal{}

\begin{document}

\begin{frontmatter}



\title{Directly Training Temporal Spiking Neural Network with Sparse Surrogate Gradient}


\author[inst1,inst2]{Yang Li}
\ead{liyang2019@ia.ac.cn}
\author[inst1,inst2]{Feifei Zhao}
\ead{zhaofeifei2014@ia.ac.cn}
\author[inst1]{Dongcheng Zhao}
\ead{zhaodongcheng2016@ia.ac.cn}
\author[inst1,inst2,inst3,inst4]{Yi Zeng\corref{cor1}}
\ead{yi.zeng@ia.ac.cn}
\cortext[cor1]{Corresponding author}

\affiliation[inst1]{organization={Brain-inspired Cognitive Intelligence Lab, Institute of Automation, Chinese Academy of Sciences (CAS)},
            city={Beijing},
            country={China}}
            
\affiliation[inst2]{organization={School of Artificial Intelligence, University of Chinese Academy of Sciences},
            city={Beijing},
            country={China}}

\affiliation[inst3]{organization={School of Future Technology, University of Chinese Academy of Sciences},
            city={Beijing},
            country={China}}

\affiliation[inst4]{organization={Center for Excellence in Brain Science and Intelligence Technology, CAS},
            city={Shanghai},
            country={China}}

\begin{abstract}
    Brain-inspired Spiking Neural Networks (SNNs) have attracted much attention due to their event-based computing and energy-efficient features. However, the spiking all-or-none nature has prevented direct training of SNNs for various applications. The surrogate gradient (SG) algorithm has recently enabled spiking neural networks to shine in neuromorphic hardware. However, introducing surrogate gradients has caused SNNs to lose their original sparsity, thus leading to the potential performance loss. In this paper, we first analyze the current problem of direct training using SGs and then propose Masked Surrogate Gradients (MSGs) to balance the effectiveness of training and the sparseness of the gradient, thereby improving the generalization ability of SNNs.
    Moreover, we introduce a temporally weighted output (TWO) method to decode the network output, reinforcing the importance of correct timesteps. Extensive experiments on diverse network structures and datasets show that training with MSG and TWO surpasses the SOTA technique. 
\end{abstract}




\end{frontmatter}

\section{Introduction}
    
    Although artificial intelligence (AI), epitomized by deep learning, has either matched or surpassed human performance in many domains, it demands substantial energy consumption. Typically, a high-end computer can consume power on the scale of megawatts or more, contrasting sharply with the human brain's mere 20 W. Such significant energy requirements hinder AI's potential for miniaturization. Brain-inspired spiking neural networks (SNNs), recognized as the third generation of artificial neural networks (ANNs) \cite{maass1997networks}, are noted for their low power consumption and rapid inference, especially when integrated into neuromorphic hardware like TrueNorth \cite{akopyan2015truenorth} and Loihi \cite{davies2018loihi}. Hybrid architectures, exemplified by Tianjic \cite{pei2019towards}, also underscore the advantages of SNNs in tandem with conventional ANNs. These benefits are primarily attributed to event-driven computation and a pronounced biological resemblance \cite{roy2019towards,zhao2022nature,zeng2022braincog}. The binary nature of these spikes means that only spiking neurons contribute to the synaptic currents, thereby minimizing redundant neural computations. Yet, this biomimetic design introduces formidable challenges in efficiently training SNNs. As such, devising effective training strategies for SNNs remains a pivotal concern, particularly for their broader practical applications.

\begin{figure}[t]
        \centering
        \includegraphics[scale=0.42]{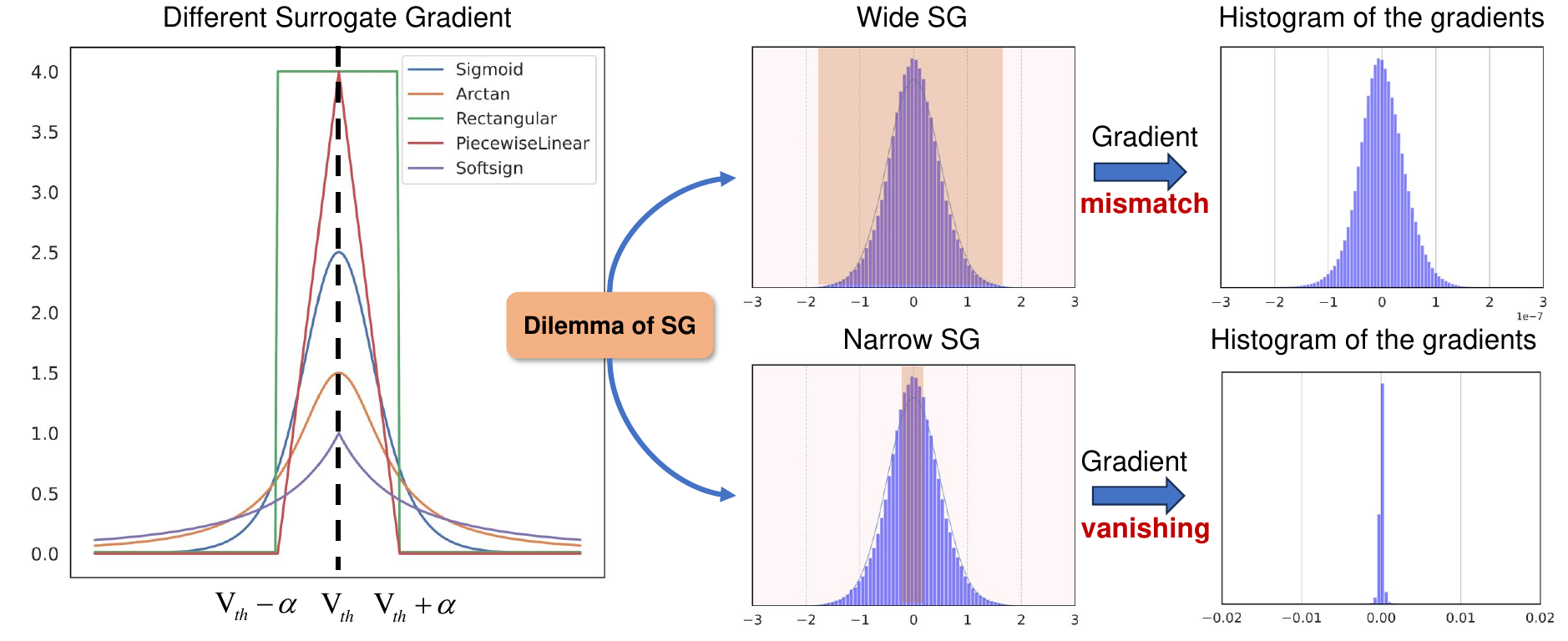}
        \caption{{Extreme surrogate gradients can result in gradient mismatch or vanishing issues. For inputs sharing a consistent membrane potential distribution, the gradient's statistical histogram reflects two polar outcomes when using either excessively wide or narrow widths: a significant portion of parameters undergo updates, or conversely, only a minimal set of parameters are updated.}}
        \label{moti}
    \end{figure}

    Recently, numerous studies have converted well-trained ANNs to SNNs \cite{diehl2015fast,han2020rmp,deng2021optimal,li2021bsnn}, aiming to harness the strengths of both backpropagation and discrete spikes. Results comparable to those of ANNs have been achieved across a range of disciplines, including image classification, target detection \cite{kim2020spiking}, semantic segmentation \cite{li2022spike}, and target tracking \cite{luo2022conversion}. However, these conversions often lead to notable performance deterioration and time delays, diminishing the rapid inference benefits associated with SNNs. Consequently, directly training SNNs to attain optimal performance has emerged as a crucial strategy for capitalizing on the full potential of SNNs, particularly within the context of neuromorphic data. 
    For the direct training of spiking neural networks, softened functions, called surrogate gradient (SG) functions \cite{wu2018spatio}, are adopted by researchers to replace the Dirac delta functions used in the original gradient, as shown in Fig. \ref{moti}, which facilitates the execution of RNN-like backpropagation through time (BPTT) algorithms. 
    While the diverse heuristic selection of surrogate gradients \cite{fang2021incorporating,bellec2018long} ensures the versatility of SNNs across various tasks, determining how the shape of these surrogate gradients aligns with specific tasks remains a significant challenge in SNN training. Given that their update rules diverge from those of ANNs, relying rigidly on particular empirical selections may hinder SNNs from identifying the most favorable solutions.

    Relative to the gradient of the original spiking process, the surrogate gradient exhibits a pronounced value augmentation at non-zero locations. As shown in Fig. \ref{moti}, expansive coverage of the surrogate gradient can result in gradient mismatch that may disrupt the training process. On the other hand, a pronounced sparsity risks the onset of gradient vanishing.
    Methods anchored in neural network searches \cite{che2022differentiable} or those adopting fixed-rate parameter optimization \cite{li2021differentiable} impose substantial computational burdens. Notably, an exclusive reliance on surrogate gradients overlooks the benefits of training derived from the pronounced sparsity of the primary gradients. A pragmatic approach entails methodically reducing the disparity between the Dirac delta function and the SG. In the backpropagation process, the SG's morphology is progressively modified to converge to or near as sharp as the Dirac delta function \cite{chen2022gradual}, promoting efficacious parameter adjustments via the surrogate gradient function during the initial training phases. This strategy effectively mitigates disturbances from incoherent gradients arising from the surrogate gradient in subsequent training stages. However, the precision required in the approximation process is non-trivial, and the potency of gradient updates often diminishes as training progresses. Dedicated endeavors are imperative to amplify the surrogate gradient's update efficiency, curtail potential disruptions during the update sequence, and bolster the network's instructional efficacy in its latter training junctures.

    To address the challenges outlined earlier, we optimize SNNs from two perspectives: sparsity and temporality, respectively. For sparsity, we propose the Masked Surrogate Gradient (MSG) method. During training, only a specific subset of parameters is updated via Stochastic Gradient Descent, while the rest are updated with zeros to maintain sparsity characteristics. Unlike the evolving surrogate gradient (ESG), MSG maintains the same surrogate gradient setting across all training phases, balancing the effectiveness of traditional surrogate gradients with the sparsity of original spiking neural network gradients. In its updates via the surrogate gradient, MSG considers the gradient sparsity inherent to the SNN, thus offering potent regularization for SNN optimization. This strategy eliminates the potential dilemma of network optimization being trapped in local minima due to an exclusive dependence on surrogate gradients. Given the limitations of surrogate gradients for optimization, the temporal dependencies intrinsic to SNNs are not accurately captured. This oversight can lead to disparities in how each timestep's output contributes to the final result. Temporally, we unveil the Temporal Weighted Output (TWO) method in a departure from gradient descent-based tweaks. This approach leverages historical statistical information to capture the importance of time-based outputs, enabling a more precise decoding. In summary, our key contributions are:
    
    \begin{itemize}
        \item We fully integrate the advantages of SG's smoothness and the sparsity of Dirac delta gradient, enabling SNNs to find the optimal path efficiently and improving the performance of SNNs.

    	\item We endow the temporal information of SNNs with adaptive weighting to precisely decode the output of SNNs, realizing that SNNs powered by dynamic temporal information achieve higher performance with fewer timesteps.
     
        \item We evaluate the proposed algorithm on multiple static and neuromorphic datasets with diverse network structures. The experimental results demonstrate that SNNs optimizing with our proposed 
        MSG and TWO methods achieve the SOTA performance. 
        For instance, we achieved top-1 accuracy rates of 95.65\% on CIFAR10 and 84.45\% on DVS-CIFAR10, respectively.
    \end{itemize}

\section{Related Work}

    \paragraph{ANN-SNN conversion} Obtaining a high-performance SNN usually consists of two methods: (1) ANN-SNN conversion and (2) direct training through the surrogate gradient. The conversion method usually maps a well-trained ANN's weights to an SNN with the same structure. It enables the SNN to perform comparably to an ANN by adjusting the weight parameters or processing the SNN information transfer. The conversion principle is to approximate the activation value of the neurons in the ANN using the firing rate of IF neurons. Diehl et al. \cite{diehl2015fast} suggested that over- or under-activation of SNNs is the cause of performance loss and propose weight normalization and threshold balancing methods. Rueckauer et al. \cite{rueckauer2017conversion} analyzed the information transfer of the conversion process in detail and suggested that the normalization parameter should be set as a percentage of the maximum activation value to achieve a more robust conversion. Han et al. \cite{han2020rmp} proposed a soft reset, in which the neuron's membrane potential is subtracted from the threshold value instead of resetting to the initial value after delivering the spikes. 
    Li et al. \cite{li2022efficient} proposed utilizing the Burst mechanism for enhanced information transmission, enabling them to achieve 95.58\% accuracy on the CIFAR-10 dataset with just 32 time steps. To further reduce the time delay, Bu et al. \cite{bu2023optimal} discovered that quantizing the activation functions of ANNs based on the characteristics of SNN information representation before conversion makes the transition smoother, thus achieving 93.96\% performance in just 4 time steps. Recently, much work has focused on implementing more complex tasks. By calibrating the spikes, Li et al. \cite{li2022spike} achieved more accurate fits in target detection and semantic segmentation tasks. Tan et al. \cite{tan2021strategy} applied conversion methods to reinforcement learning tasks to achieve results comparable to ANNs. In conclusion, although the conversion method enables SNNs to perform like ANNs in various tasks, the considerable time delay still needs to be addressed urgently.

    \paragraph{Direct training through surrogate gradient} Training spiking neural networks with the surrogate gradient method can avoid training crashes or inefficiencies \cite{zenke2018superspike}. It fully draws on the theoretical advantages of the BP to utilize the global spatiotemporal information of the SNN and thus achieve the desired accuracy and low inference time delay.
    STBP \cite{wu2018spatio} assigned credit to the spatiotemporal information using BPTT. Due to the full use of the spatiotemporal representation, the deep spiking neural network performance is significantly improved. Rectangular SG is used in STBP. Then researchers have successively proposed various shapes of surrogate gradients, such as Sigmoid functions \cite{bengio2013estimating}, Arctan functions \cite{fang2021incorporating}, piecewise linear functions \cite{bellec2018long}, and piecewise exponential functions \cite{shrestha2018slayer}. TET \cite{deng2022temporal} achieves the performance of 74.47\% and 83.17\% on datasets like CIFAR-100 and DVS-CIFAR10, respectively, by directly calculating the loss at each output moment. Differentiable Spike \cite{li2021differentiable} theoretically investigated the gradient descent problem of training in SNNs and introduced effective differential gradients to analyze the training behavior of SNNs quantitatively. They proposed a set of differentiable spiking functions based on the introduced finite differential gradients. Dspike can evolve adaptively during training to find the best shape and smoothness of the gradient. Recently, researchers designed a progressive surrogate gradient learning algorithm \cite{chen2022gradual} to ensure that the spiking neural network effectively back-propagates gradient information early in training and obtains more accurate gradient information later.

    \paragraph{Spike Decoding} Due to the limitations of spike representation compared to real-value representation, many studies attempt to utilize the temporal characteristics of SNNs to express more information. Techniques like Burst~\cite{li2022efficient} and DSR \cite{meng2022training} try to enhance the informational capacity of spikes by weighting spike sequences with fixed weights \cite{xiao2021training}, achieving performance comparable to ANNs with fewer time steps. Some work \cite{stockl2021optimized} also attempts to represent richer information with fewer spikes in a learnable manner, further exploiting the intervals between spikes to achieve greater sparsity, which often relates to the computational energy consumption of SNNs. These works all emphasize the importance of enriching the representation of neuronal information across the entire SNN network.
    Similar to the studies above, TWO also leverages the temporal characteristics of SNNs but focuses only on the output layer since it is more closely related to the supervised labels. The weighting coefficients integrate the advantages of fixed weights, which do not increase the optimization burden, and the higher flexibility of learnable weights, slidingly changing their coefficients according to the results of supervised learning during training and remaining fixed during inference. Thus, TWO focuses on amplifying the impact of temporal components contributing more to correct classification in past learning processes, directly affecting the network's classification performance.

\section{Preliminaries}

    In this section, we introduce the LIF model and the surrogate gradient used in the proposed method theoretically and describe the encoding and decoding of SNNs.

    \subsection{Leaky Integrate-and-Fire Model}
    The LIF is the most commonly used neuron model in SNN, where a neuron receives spikes from a presynaptic neuron and dynamically changes the membrane potential. The updating process can be described as follows:
    \begin{align}
    	\tau \frac{d\textbf{V}^{(l)}(t)}{dt} = - (\textbf{V}^{(l)}(t)-\textbf{V}_{reset}) + \textbf{I}^{(l)}(t)
    \end{align}
    where $\tau$ is the time constant, $l$ and $t$ mean the indexes of layer and timestep, $\textbf{V}^{(l)}$ denotes the membrane potential, and $I$ represents the current input from the presynaptic neuron and is the inner product of synaptic weights $\textbf{W}^{(l)}$ and spiking inputs of last layer $\textbf{S}^{(l-1)}$, i.e., $\textbf{I}^{(l)}=\textbf{W}^{(l)}\textbf{S}^{(l-1)}$.
    Given a neuronal threshold $V_{th}$, when the neuron does not emit spikes ($V^l<V_{th}$), its membrane potential gradually leaks to the resting potential $\textbf{V}_{rest}$. Usually, $\textbf{V}_{rest}=\textbf{V}_{reset}=0$. When the neuron's membrane potential exceeds the threshold $V_{th}$, the neuron delivers a spike, and the potential resets to the resting potential $V_{rest}$. For computational tractability, we simplify the equation to an iterative form, as shown in the following equation:
    \begin{gather}
    \textbf{I}^{(l)}[t+1]=\textbf{W}^{(l)}\textbf{S}^{(l-1)}[t+1]\\
    	\textbf{V}^{(l)}[t+1] = \textbf{V}^{(l)}[t] - \frac{1}{\tau}(\textbf{I}^{(l)}[t+1]-\textbf{V}^{(l)}[t])\\
    	\textbf{S}^{(l)}[t+1] = \Theta(\textbf{V}^{(l)}[t+1]-\textbf{V}_{th})\\
    	\textbf{V}^{(l)}[t+1] = \textbf{V}^{(l)}[t+1](1-\textbf{S}^{(l)}[t+1])
    \end{gather}
    where $\Theta$ symbolizes the Heaviside step function. The output spike will be passed as the presynaptic spike of the next layer, thus completing the information transfer. In this paper, we set the threshold $\textbf{V}_{th}$ to 0.5 and the time constant $\tau$ to 2.

    In addition, the encoding of the input and the decoding of the output loses the exact information of the image and the output potential information to some extent, so we take the real value $\textbf{X}$ directly for the input and encode the input information to spike trains $\textbf{S}^{(1)}$ by the first convolutional and LIF layer. For the network's output, we integrate the synaptic current of the output layer without letting it leak over time or emit spikes. Finally, we set the mean value of the total synaptic current as the network's output.

    \subsection{Surrogate Gradient of SNN}
    For directly training the SNNs, given a loss function $\mathcal{L}$ for a task, the SNN update process can be expressed according to the chain rule as
    \begin{align}
    	\frac{\partial \mathcal{L}(\textbf{W})}{\partial \textbf{W}}=\sum\limits _t \frac{\partial \mathcal{L(\textbf{W})}}{\partial \textbf{S}[t]} \frac{\partial \textbf{S}[t]}{\partial \textbf{V}[t]} \frac{\partial \textbf{V}[t]}{\partial \textbf{I}[t]} \frac{\partial \textbf{I}[t]}{\partial \textbf{W}}
    	\label{eq_loss}
    \end{align}
    where $\frac{\partial \textbf{S}[t]}{\partial \textbf{V}[t]}$ is the gradient of the spiking function, which is identified as the Dirac delta function. It is not a good choice for gradient descent-based learning because it is 0 everywhere except at $V[t] = V_{th}$. To solve the problem, the surrogate gradient uses a smooth function instead of the original gradient, and a typical family is the Arctan function\cite{fang2021incorporating}.
    \begin{align}
    	\frac{\partial \textbf{S}[t]}{\partial \textbf{V}[t]} = \frac{\alpha}{1 + (\frac{\pi}{2}\alpha (\textbf{V}[t]-\textbf{V}_{th}))^2}
    \end{align}
    The surrogate gradient is calculated based on the distance of the membrane potential $\textbf{V}[t]$ from the threshold $\textbf{V}_{th}$, where $\alpha$ is the constant factor which determines the shape of the surrogate gradient. In this paper, we also use another type of SG Piecewise Linear Grad (PLGrad) for comparison:
    \begin{align}
    	\frac{\partial \textbf{S}[t]}{\partial \textbf{V}[t]} = \operatorname{max}\left(0, \alpha \left(1-\alpha|\textbf{V}[t]-\textbf{V}_{th}|\right)\right)
    \end{align}
    Then, we can update the gradient like a traditional BP:
    \begin{align}
    	\textbf{W} \leftarrow \textbf{W}- \eta \frac{\partial \mathcal{L}(\textbf{W})}{\partial \textbf{W}}
    \end{align} 

    The surrogate gradient technique leverages only the information derived from the SNN during forward propagation, omitting the conventional backpropagation data. As depicted in Eq. \ref{eq_loss}, the term $\frac{\partial \textbf{I}[t]}{\partial \textbf{W}}$ can be condensed to the pre-synaptic spike, denoted as $\textbf{S}[t]$. When paired with this value, a smooth gradient facilitates the parameter updates, adeptly avoiding the pronounced sparsity characteristic of the original gradient.

    \section{Methodology}
        \subsection{Rethink the Surrogate Gradient}

    As illustrated in Fig. \ref{moti}, multiple surrogate gradient variants are used to train SNNs directly. The membrane potential range they address directly influences the forward propagation depth of the gradients. A broader surrogate gradient covers a more extensive membrane potential input range, facilitating more pronounced gradient updates. Nonetheless, this might result in undesirable updates that can hinder the network's pursuit of optimal solutions, giving rise to gradient adaptation challenges. On the other hand, a narrower surrogate gradient more closely mirrors the Dirac delta gradient's contour, more accurately reflecting the original gradient update direction. The heightened sparsity of the gradient often curtails further forward propagation, which can lead to gradient vanishing dilemmas. 
    Methods such as ESG, which employ crisper surrogate gradients in the later stages of training, address gradient adaptation to a degree but risk undermining the network's learning prowess. This assertion is supported by multiple learning rate decay strategies \cite{loshchilov2016sgdr}, which show that networks exhibit strong learning potential even in advanced training stages. Therefore, making full use of the surrogate gradient's learning ability and maintaining the gradient's sparsity during direct training is a problem that needs to be solved.

    \begin{figure*}[!t]
    	\centering
    	\includegraphics[scale=0.35]{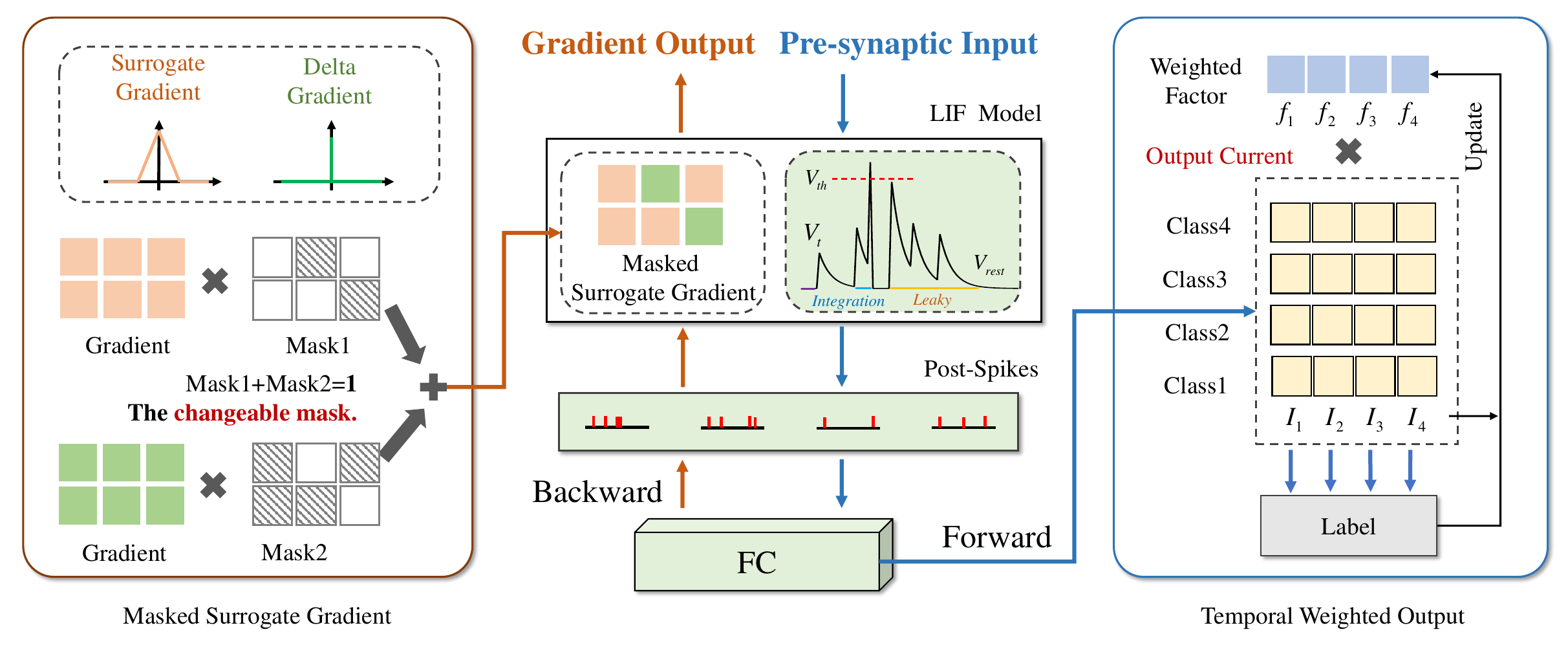}
    	\caption{\textbf{The whole workflow of the proposed methods.} MSG generates a random mask before calculating the gradient, then uses smooth SG to assist training, and mask, on the other hand, provides stronger sparsity to mitigate the interference of surrogate gradients in training. TWO updates the weighted factors based on the historical output correctness of each timestep, reinforcing the importance of those moments that can be classified correctly in a single timestep.}
    	\label{whole}
    \end{figure*}

\subsection{Masked Surrogate Gradient}
    \label{msg}

    The Masked Surrogate Gradient (MSG) is motivated by the idea of using surrogate gradients to supplement, rather than supplant, the updates of SNNs, for efficient and high-performance SNN direct optimization. This approach aims to harness both the surrogate gradient's robust updating potential and the original gradient's pronounced sparsity throughout the training process, as illustrated in Fig. \ref{whole}. Define $\mathcal{G}=\frac{\partial \mathcal{L}(\textbf{W})}{\partial \textbf{W}}$, and $\mathcal{G}_{SG}$ uses surrogate function to compute $\frac{\partial \textbf{S}[t]}{\partial \textbf{V}[t]}$. Like traditional gradient descent, the MSG computes gradients for all trainable parameters. The difference, however, is that MSG uses a surrogate for only some of the parameter updates at $t$ iterations, whereas the remainder does not update. The MSG offers potent parameter updating capabilities coupled with heightened sparsity. Also, it always concerns the network's potential inefficacy to converge in initial training stages or its waning update prowess in subsequent stages. Initially, we define a binary mask, congruent in shape to \textbf{W}, and generate it following a Bernoulli distribution characterized by probability $p$. When $p=0$, MSG degenerates to the conventional SG approach.
    
    \begin{align}
    	\mathcal{M}_b \sim Bernoulli(1-p)
    \end{align}
    Then, we define the update process of the weights as shown in the following equation. The higher the probability of $p$, the smaller the fraction of the surrogate.
    \begin{gather}
    	\mathcal{G} = \mathcal{G}_{SG} \odot \mathcal{M}_b \\
    	\textbf{W} \leftarrow W\textbf{}-\eta \mathcal{G}
    	\label{up}
    \end{gather}

    To make each neuron have the same chance of being masked, we randomly generate a mask based on the mask probability for each neuron layer under each minibatch under each epoch when calculating the gradient during feedback, such that some of the gradients obtained by using the surrogate gradient function are computed to be zero. The MSG adds only one step of masking to the original gradient and thus does not significantly increase the simulation's runtime of the simulation. Then, we discuss how MSG helps SNNs better find optimal solutions. Due to the employment of the mask, the derived gradient values are predominantly zero, especially when contrasted with the smooth SG function. Although momentum introduced by the optimizer ensures parameter updates aren't exactly zero, the scope of MSG updates remains notably less extensive than those of SG. We find the sparsity introduced by MSG subsequently amplifies the standard deviation of the gradient. To elucidate this point, consider the following:
    
    Suppose the gradient update obeys Gaussian distribution $\overline{\mathcal{G}}_i \sim \mathcal{N}(\mu, \sigma^2)$, if the gradient update is masked with probability $p$ (the mask value $m=0$), suppose the update is 0 when input $x=0$, then the mean and variance of the gradient update are
    $\mathbb{E}[\mathcal{G}_i]=\mu$,
    $Var[\mathcal{G}_i]=\sigma^2+\frac{p}{1-p}\left(\mu^2+\sigma^2\right)$.

    Suppose $\mathcal{G}_i=\overline{\mathcal{G}}_i\frac{m}{1-p}$, then we can get
    \begin{align}
        \mathbb{E}[\mathcal{G}_i]=\mathbb{E}\left[\overline{\mathcal{G}}_i\frac{m}{1-p}\right]=\mathbb{E}[m]\mathbb{E}[\overline{\mathcal{G}}_i]/(1-p) = \mu
    \end{align}
    Because $Var[\mathcal{G}_i]=\mathbb{E}[\mathcal{G}_i^2]-\mathbb{E}[\mathcal{G}_i]^2$, therefore,
    \begin{align}
        \begin{split}
            Var[\mathcal{G}_i] &= \mathbb{E}[\overline{\mathcal{G}}_i^2]\frac{1-p}{(1-p)^2}-\mathbb{E}[\mathcal{G}_i]^2 \\
            &= \frac{\mu^2 + \sigma^2}{1-p}-\mu^2\\
            &= \sigma^2 + \frac{p}{1-p}(\mu^2+\sigma^2)
        \end{split}
        \label{var}
    \end{align}
    
    We can see from Eq. \ref{var} that the variance is a strictly increasing function of $p$. Thus, MSG can be seen as a strong regularization of the SNN training process. It allows the SNN to skip the saddle points of the loss landscape and find the local minima that fit itself.

\subsection{Temporally Weighted Ouput}

    SNNs exhibit the advantage of low energy consumption attributed to their sparsity and also in their integration of temporal and spatial information. Motivated by this, we combine MSG with Temporally Weighted Output (TWO) to obtain more accurate inference for SNNs.
    During the update process, a disparity persists between the loss landscape of SG and the actual one. As a result, when utilizing the BPTT algorithm, SG encounters difficulties in precisely adapting to the information dependencies at each time step. We collected data on the output accuracy at each time step over the entire training process and noticed that, with training progression, the later time steps often demonstrated improved performance. Building on this observation, we introduced a Temporally Weighted Output approach to enhance the decoding of the network's outputs, integrating a temporal importance factor (TIF). Initially, we set the factor $f_t=\frac{1}{T}$  for each timestep and then the output for each timestep as
    \begin{align}
    	\textbf{y}[t] = f_t \textbf{I}^{(L)}[t]
    \end{align}
    where $L$ means the index of the last model layer. One possible approach is to learn it as a trainable parameter. However, this increases the burden of gradient descent to find the optimal point, especially when $T$ is large. Therefore, we count the historical output's accuracy to determine each timestep output's weight.
    \begin{align}
    	\label{df}
    	\Delta_{f_t} = \frac{\sum\limits_{i}^{N}(y_i[t]==\hat{y_i})}{N |\textbf{y}|}\\
    	f_t \leftarrow \beta f_t + (1-\beta) \Delta_{f_t}
    	\label{ft}
    \end{align}

    \begin{figure}[t]
    	\centering
    	\includegraphics[scale=0.5]{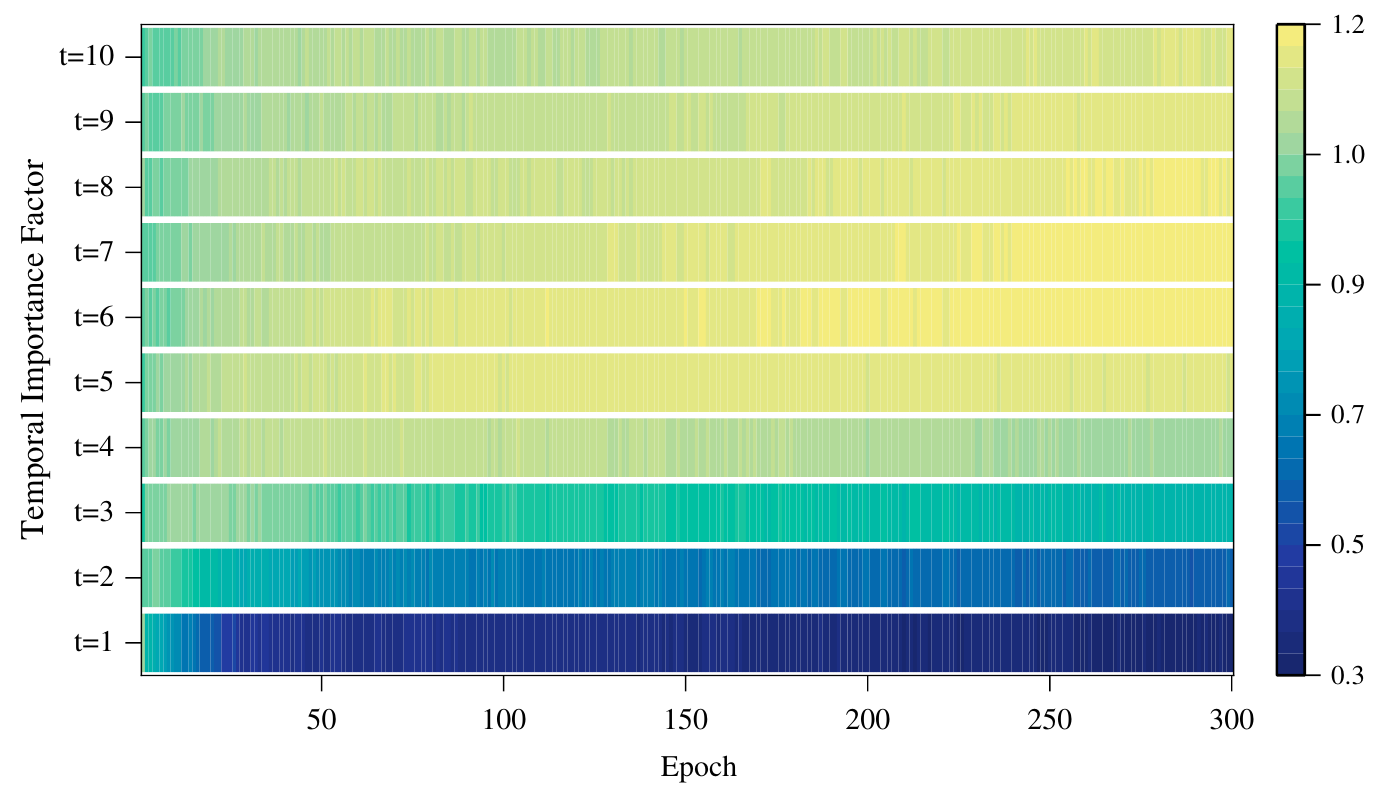}
    	\caption{{The changes of the temporal importance factor varies with epoch. While an identical factor is assigned at the onset, the training progression reveals a discernible trend: the accuracy at initial time steps is frequently suboptimal compared to the elevated accuracy observed in subsequent steps. Consequently, this temporal importance factor eventually stabilizes to a revised value, enhancing the output decoding efficacy of SNN.}}
    	\label{factor}
    \end{figure}

    \begin{algorithm}[!b]
    	\caption{Directly Training SNNs with MSG and TWO.}
    	\label{alg2}
    	\textbf{Input}: SNN to be trained; Training dataset; Testing dataset; Total timestep $T$; Total training epochs $E$; Initialized weighted fator $f=[\frac{1}{T}, \frac{1}{T}, \cdots, \frac{1}{T}]^{\top}$; Mask probability $p$; The parameter of surrogate gradient.\\
    	\textbf{Output}: The well-trained SNN.
    	\begin{algorithmic}[1] 
    		\FOR{all $i$=1, 2, 3, $\cdots$, $E$ epoch}
    		\STATE Get input data and class label: $\hat{\textbf{y}_i}$;\\
    		\STATE Calculate the output: $\textbf{y}_i = \frac{1}{T}\sum\limits_{t}^T f_t \textbf{I}^{(L)}[t]$;\\
    		\STATE Calculate the update of the weighted factor $\Delta_{f_t}$ according to Eq. (\ref{df}) and (\ref{ft}) ;\\
    		\STATE Compute the total loss;\\
    		\FOR{all $l=L, L-1, \cdots, 1$ layer}
    		\STATE Calculate the gradients $\mathcal{G}_{SG}$;\\
    		\STATE Generate the mask $\mathcal{M}_b$ with probability $p$;\\
    		\STATE Calculate new gradients $\mathcal{G}_{SG} \odot \mathcal{M}_b$;\\
    		\STATE Update model parameters according to Eq. (\ref{up});\\
    		\ENDFOR
    		\ENDFOR
    	\end{algorithmic}
    \end{algorithm}

    Then, we use low-pass filtering for its update process to keep it stable. $ \beta$ is the constant. Thus, a better decoding of the output information is achieved. In Fig. \ref{factor}, we depict the evolution of the temporal importance factor for each timestep over successive epochs. While the factor commences identically, the training course unveils an inferior performance of the earlier timesteps, causing their factor to diminish persistently.
    The feedforward and backward process training algorithm is detailed in Algorithm \ref{alg2}.

\section{Experiments}

In this section, we validate the proposed MSG and TWO algorithms and compare them with previous works on static and neuromorphic datasets. In addition, adequate ablation experiments are used to verify the effectiveness of the proposed methods. For the static dataset, we encode real-valued inputs to spike using the first layer of the network and use the CIFARNet structure (256C3-256C3-AP2- 256C3-256C3-AP2- 512C3-512C3- 512C3-AP2- 1024C3- 1024C3-AP2-FC). For the neuromorphic dataset, we output it as a fixed number of frames as in the previous works. For comparison, we follow \cite{deng2022temporal} to use the VGGNet structure (64C3-128C3-AP2-256C3-256C3-AP2-AP2- 512C3-512C3-AP2- 512C3-AP2-FC), which is a variant of the VGG11 network with only one fully connected layer. In addition, ResNet18 is also used for the neuromorphic dataset.

\subsection{Comparison to Previous Works}

\begin{table*}[!t]
	\centering
 \caption{Performance comparisons of the proposed method and previous work.}
	\resizebox{\columnwidth}{!}{    
		\begin{tabular}{lllrrc}
			\toprule
			Dataset  &  Method & Type &Architecture & Timestep & Accuracy (\%)\\
			\midrule
			\multirow{9}{*}{CIFAR10} 
			&  Burst Spike \cite{li2022efficient} & ANN-SNN Conversion  & VGG16 & 32 & \textbf{95.58} \\
			&  ACP \cite{li2022converting} & ANN-SNN Conversion  & VGG16 & 32 & 94.81 \\
			&  Dspike \cite{li2021differentiable} & Direct training  & ResNet18 & 4 & 93.66$\pm$0.05 \\
			&  TET \cite{deng2022temporal} & Direct training  & ResNet19 & 4 & 94.44$\pm$0.08 \\
			&  tdBN \cite{zheng2021going} & Direct training  & ResNet19 & 4 & 92.92\\
			&  TSSL-BP \cite{zhang2019spike} & Direct training  & CIFARNet & 5 & 91.41 \\
			&  Diet-SNN \cite{rathi2020diet} & Direct training  & CIFARNet & 5 & 91.59 \\
			&  RecDis-SNN \cite{guo2022recdis} & Direct training  & CIFARNet & 4 & 92.20$\pm$0.10 \\
			\cmidrule{2-6}

            & Ours (w/o AutoAugment)  &  \multirow{2}{*}{{Direct training}}
			&\multirow{2}{*}{CIFARNet} & \multirow{2}{*}{4} & 94.30$\pm$0.22 \\
            & Ours (w/ AutoAugment) & & & & 95.40$\pm$0.12\\
			\midrule
			
			\multirow{9}{*}{CIFAR100} 
			&  Burst Spike \cite{li2022efficient} & ANN-SNN Conversion  & VGG16 & 32 & 74.98 \\
			&  ACP \cite{li2022converting} & ANN-SNN Conversion  & VGG16 & 32 & 75.53 \\
			&  GSG \cite{chen2022gradual} & Direct training  & VGG-small & 5 & 69.20$\pm$0.19 \\
			&  Dspike \cite{li2021differentiable} & Direct training  & ResNet18 & 4 & 73.35$\pm$0.14 \\
			&  TET \cite{deng2022temporal} & Direct training  & ResNet19 & 4 & 74.47$\pm$0.15 \\
			&  RecDis-SNN \cite{guo2022recdis} & Direct training  & VGG16 & 5 & 69.88$\pm$ 0.08 \\
			\cmidrule{2-6}

                & Ours (w/o AutoAugment)  &  \multirow{2}{*}{{Direct training}}
			&\multirow{2}{*}{CIFARNet} & \multirow{2}{*}{4} & 77.41$\pm$0.07 \\
            & \textbf{Ours (w/ AutoAugment)} & & & & \textbf{77.48$\pm$0.14}\\
			\midrule

                \multirow{5}{*}{ImageNet} 
			&  Burst Spike \cite{li2022efficient} & ANN-SNN Conversion  & VGG16 & 32 & \textbf{70.61} \\
			&  ACP \cite{li2022converting} & ANN-SNN Conversion  & VGG16 & 32 & 69.04 \\
                &  TET \cite{deng2022temporal} & Direct training  & Sew-ResNet34 & 4 & 68.00 \\
			&  RecDis-SNN \cite{guo2022recdis} & Direct training  & ResNet34 & 6 & 67.33 \\
			\cmidrule{2-6}
                & Ours  &  \multirow{1}{*}{{Direct training}}
			&\multirow{1}{*}{Sew-ResNet34} & \multirow{1}{*}{4} & 67.46 \\
                \midrule
			
			\multirow{9}{*}{DVS-CIFAR10} 
			&  DART \cite{ramesh2019dart} & Direct training  & -  & - & 65.78 \\
			&  tdBN \cite{zheng2021going} & Direct training  & ResNet19 & 10 & 67.80\\
			&  NDA \cite{li2022neuromorphic} & Direct training  & VGG11  & 10 & 81.70 \\
			&  RecDis-SNN \cite{guo2022recdis} & Direct training  & CIFARNet & - & 67.30$\pm$0.05 \\
			&  Dspike \cite{li2021differentiable} & Direct training  & ResNet18 & 10 & 75.40$\pm$0.05 \\
			&  TET \cite{deng2022temporal} & Direct training  & VGGSNN & 10 & 83.17$\pm$0.15 \\
			\cmidrule{2-6}
			&\multirow{2}{*}{\textbf{Ours}} &  \multirow{2}{*}{{Direct training} }
			& {ResNet18} & {10} & {79.35$\pm$ 0.34}\\
			&&& {VGGSNN} & {10} & \textbf{83.97 $\pm$ 0.12} \\
			\midrule
			
			\multirow{5}{*}{NCALTECH101} 
			&  DART \cite{ramesh2019dart} & Direct training  & -  & - & 66.42 \\
			&  SALT \cite{kim2021optimizing} & Direct training  & -  & 20 & 55.00 \\
			&  NDA \cite{li2022neuromorphic} & Direct training  & VGG11  & 10 & \textbf{83.70} \\
			\cmidrule{2-6}
			&\multirow{2}{*}{{Ours}} &  \multirow{2}{*}{{Direct training} }
			& {ResNet18} & {10} &  {70.81$\pm$0.31}\\
			&&&{VGGSNN} & {10} &  77.70$\pm$0.15\\
			\bottomrule
	\end{tabular}}
	
	\label{all}
\end{table*}

    The performance of our proposed method is compared with some SOTA methods, as listed in Table \ref{all}. Our experiments are performed on the NVIDIA A100 based on the PyTorch framework. We use AutoAgument \cite{cubuk2018autoaugment} and Cutout \cite{devries2017improved} for data augmentation on the CIFAR dataset. For neuromorphic data, there is no data augmentation used. We use adamW\cite{loshchilov2017decoupled} as the optimizer with an initial learning rate of 0.05. A cosine learning rate decay strategy changes the learning rate dynamically. For all experiments except for ImageNet in Tab. \ref{all}, we train the SNN for 300 epochs. We set different random seeds to repeat the experiments three times to avoid random errors. We also replace all the MaxPooling layers in the network with average pooling.

    Our MSG achieves 95.40\% top-1 accuracy on CIFAR10 with four timesteps,  which significantly outperforms the RecDis-SNN \cite{guo2022recdis} with 3.20\% improvements. The reported accuracy is also better than TSSL-BP \cite{zhang2019spike} and Diet-SNN \cite{rathi2020diet} using fewer time steps. In addition, MSG uses fewer time steps to achieve comparable performance with the conversion method. 
    For CIFAR100, 77.48\% top-1 accuracy is reached, which is better than all direct training SNN methods and the conversion methods with fewer time steps. We also report results without applying autoaugment and Cutout, which show that the performance degradation of the network is low in the CIFAR dataset. To test the ability of MSG to optimize on larger data, we test the performance of Sew-ResNet34 on ImageNet, achieving a performance of 67.46\% in 4 time steps, which shows that MSG does not affect the convergence of SNNs despite providing regularization. Note that it is slightly lower than the 68\% of the TET method. We analyse this because on the ImageNet dataset, the input information is the same at each time step, so TET forces the sparse outputs of each time step to approximate towards the labels more effectively for learning.

    Unlike conversion methods, SNN can handle neuromorphic data with temporal information. DVS-CIFAR10 contains 10k images transformed from CIFAR10 and is partitioned into a training and a test set in a 9:1 ratio. Our method obtains 83.97\% top-1 accuracy in DVS-CIFAR10, nearly a 0.8\% improvement with the best TET \cite{deng2022temporal} method. NCALTECH101 has many categories and few samples, and we do not perform data augmentation on the neuromorphic data. As we all know, data augmentation is often more effective for overfitting problems. Hence, the results are slightly worse than NDA \cite{li2022neuromorphic}, but the accuracy of 77.70\% also outperforms the other methods.

	In conclusion,  MSG sparsifies the surrogate gradient, making the MSG method less affected by different surrogate gradient settings and hyperparameters, achieving better performance across multiple datasets and network configurations. Thanks to the sparse operations in MSG, we can control the gradient update process through a gating mechanism, thereby reducing the computational load of gradient updates. Particularly, in scenarios requiring online learning, the MSG method can reduce computational load while maintaining model performance, thereby improving computational efficiency.

    \begin{figure}[t]
    	\centering
    	\includegraphics[scale=0.4]{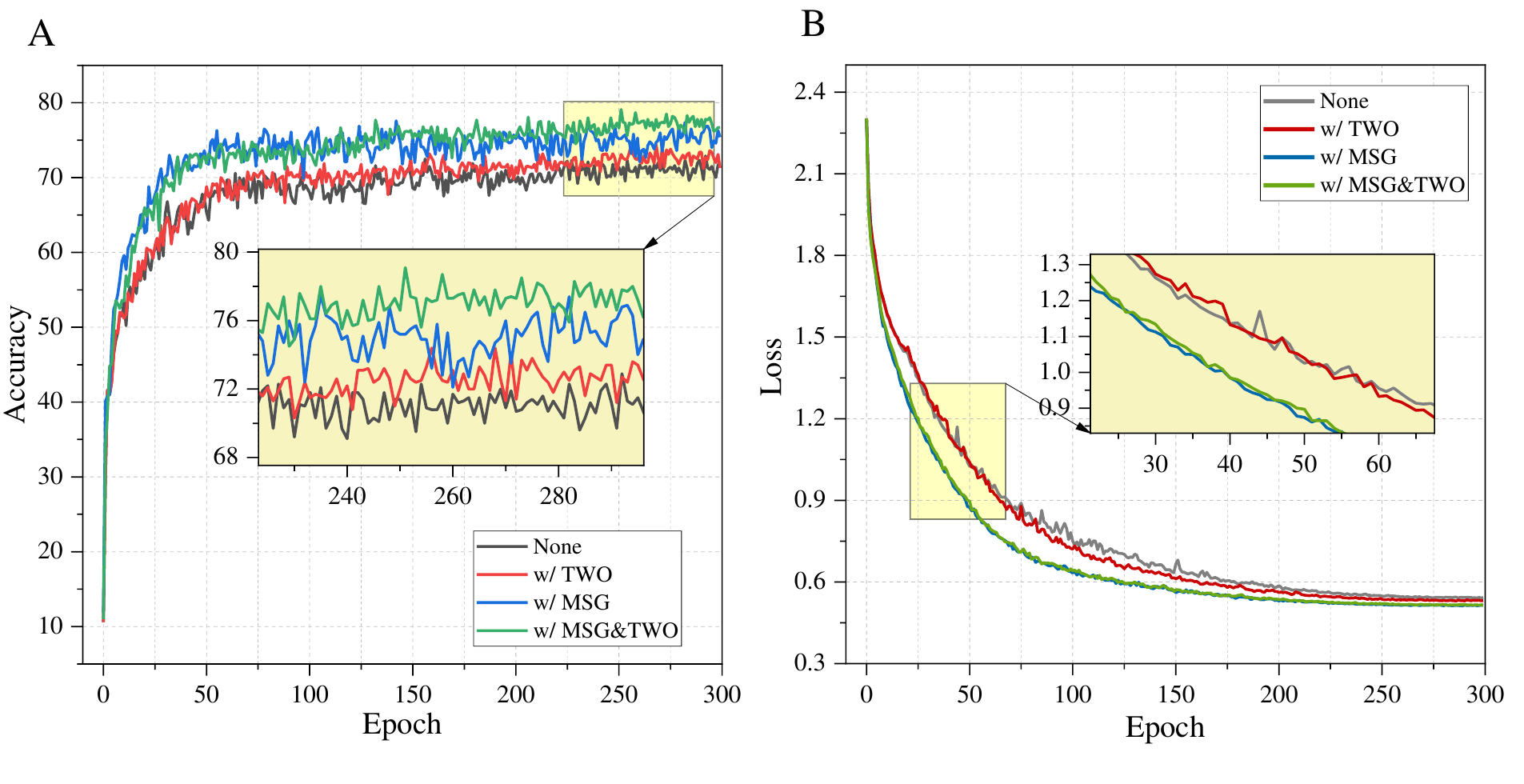}
    	\caption{\textbf{MSG and TWO help to improve the performance.} We provide  (A)  the test accuracy and (B) the training loss curve on DVS-CIFAR10 with ResNet18. MSG helps to jump out the local minimum point, and TWO improves the ability to decode the output.}
    	\label{abl}
    \end{figure}

\begin{table}[t]
	\centering
 \caption{Ablation study on CIFAR10 and DVS-CIFAR10.}
	\resizebox{0.8\columnwidth}{!}{    
		\begin{tabular}{lcclcc}
			\toprule
			Dataset & Network & SG Type &  Method & Accuracy (\%) \\
			\midrule
			\multirow{4}{*}{CIFAR100}  & \multirow{4}{*}{ResNet18} & \multirow{4}{*}{AtanGrad}
			& None & 68.79\\
			&&& w/ TWO & 68.98\\
			&&& w/ MSG & 69.08\\
			&&& w/ MSG \& TWO & \textbf{69.21}\\
			\midrule
			\multirow{4}{*}{DVS-CIFAR10}  & \multirow{4}{*}{ResNet18} & \multirow{4}{*}{AtanGrad}
			& None & 72.90\\
			&&& w/ TWO & 74.40\\
			&&& w/ MSG & 78.30\\
			&&& w/ MSG \& TWO & \textbf{79.10}\\
                \midrule
			\multirow{4}{*}{DVS-CIFAR10}  & \multirow{4}{*}{VGGSNN} & \multirow{4}{*}{PLGrad}
			& None & 82.70\\
			&&& w/ TWO & 83.20\\
			&&& w/ MSG & 83.30\\
			&&& w/ MSG \& TWO & \textbf{83.90}\\
			\bottomrule
	\end{tabular}}
	
	\label{ab}
\end{table}

\subsection{Ablation Study}

\paragraph{Effectiveness of MSG and TWO} In this section, we perform experiments to validate the proposed MSG and TWO in the ResNet18 and VGGSNN using the CIFAR100 and DVS-CIFAR10 datasets. The total timestep is 4 in the static dataset and 10 in the neuromorphic dataset, with a mask probability of 0.5. We use AtanGrad and PLGrad and train the network for 300 epochs, and the detailed results are shown in Table \ref{ab}.
As a decoding method, TWO utilizes the output correctness of individual timestep to update the weights for each moment empirically. So, it does not contribute to minimizing the loss in the optimization phase, as shown in Fig. \ref{abl}. However, we can see from Fig. \ref{abl} (A) and Table \ref{ab} that TWO can take advantage of the high-weight information in the output currents to improve the classification performance. In Table \ref{ab}, we find that TWO only improves slightly on CIFAR100 compared to the baseline. A possible explanation is that the time step of CIFAR100 is less, and the static dataset has the same input at each moment, so TWO has less space for further decoding the output. In contrast, the improvement of TWO on DVS-CIFAR10 is significant, which indicates that TWO is more advantageous on neuromorphic datasets and networks with larger timestep.

As shown in the blue and green lines in Fig. \ref{abl}, MSG can significantly minimize the loss and achieve a consistent performance boost on both datasets. The surrogate gradient, while providing a smooth gradient for the SNN, also adds a lot of interference compared to the original gradient. MSG can fully combine both gradients to find a more suitable optimizing path for the SNN to minimize the loss and improve the network's learning ability. We will provide a detailed analysis of why MSG can work later on. Finally, the combination of MSG and TWO achieves optimal accuracy of 69.21\% and 79.10\% on CIFAR100 and DVS-CIFAR10, respectively, demonstrating the effectiveness of the proposed method.

\begin{figure}[!t]
	\centering
	\includegraphics[scale=0.5]{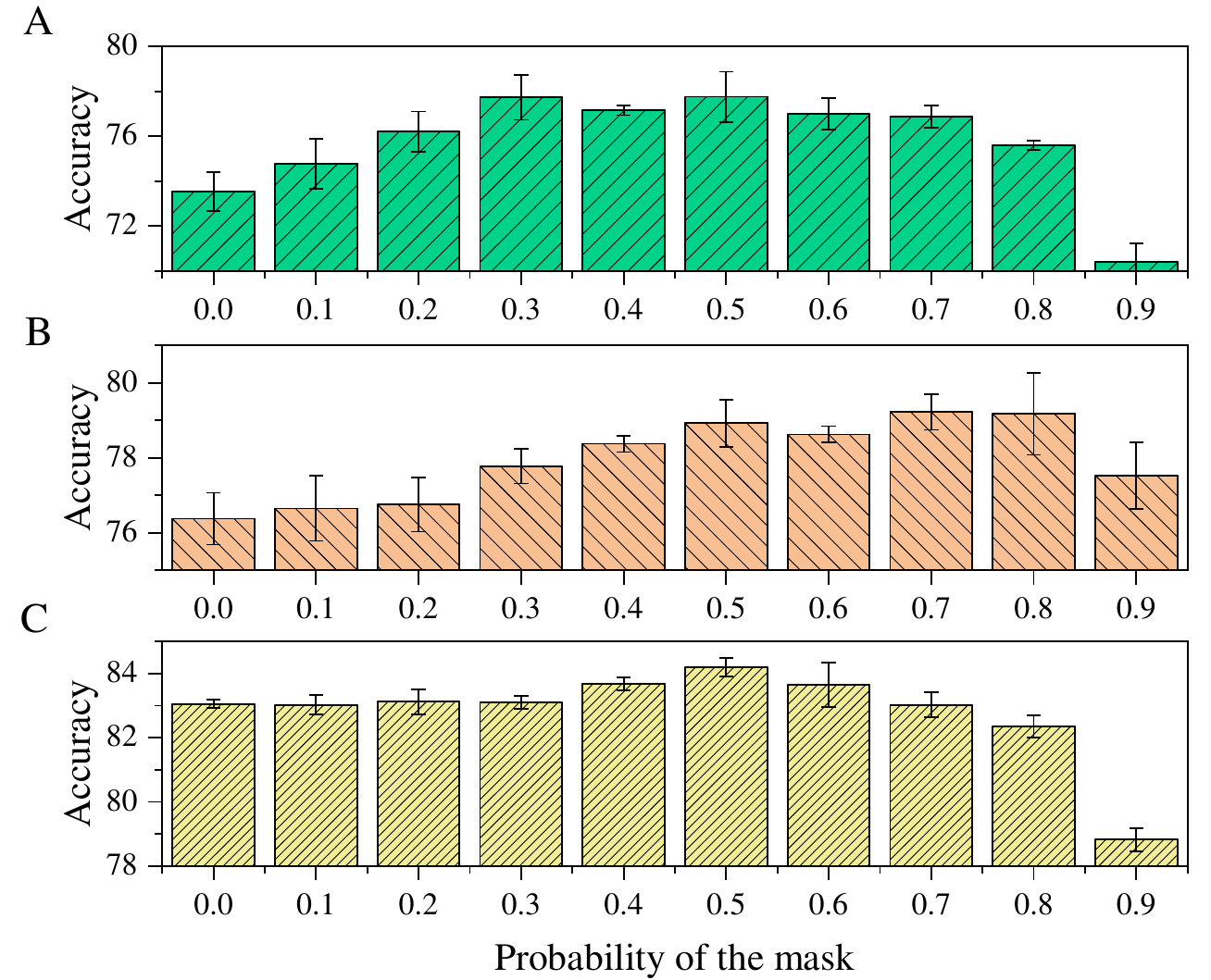}
	\caption{{\textbf{The effect of mask probability.} We test the classification performance of SNNs with different network structures and surrogate gradients on DVS-CIFAR10 under different mask probabilities. (A) ResNet18 with Arctan gradient. (B) VGGNet with Arctan gradient. (C) VGGNet with PiecewiseLinear gradient. }}
	\label{figp}
\end{figure}

\begin{figure*}[t]
    \centering
    \subfigure[Firing rate in VGGSNN.]{
        \label{sparsity}
        \includegraphics[scale=0.27]{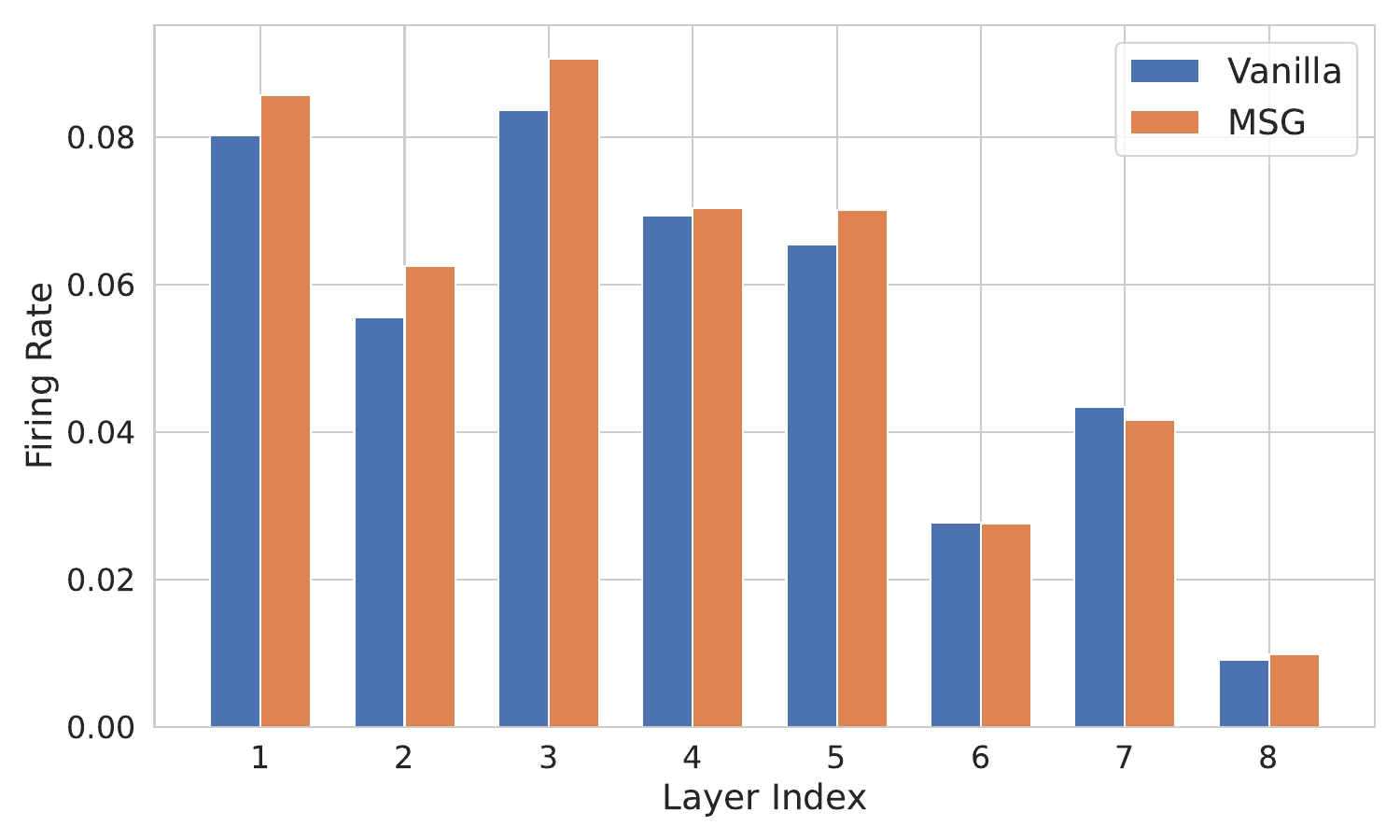}} \hspace{2pt}
    \subfigure[Training time per epoch on DVS-CIFAR10.]{
        \label{time}
        \includegraphics[scale=0.29]{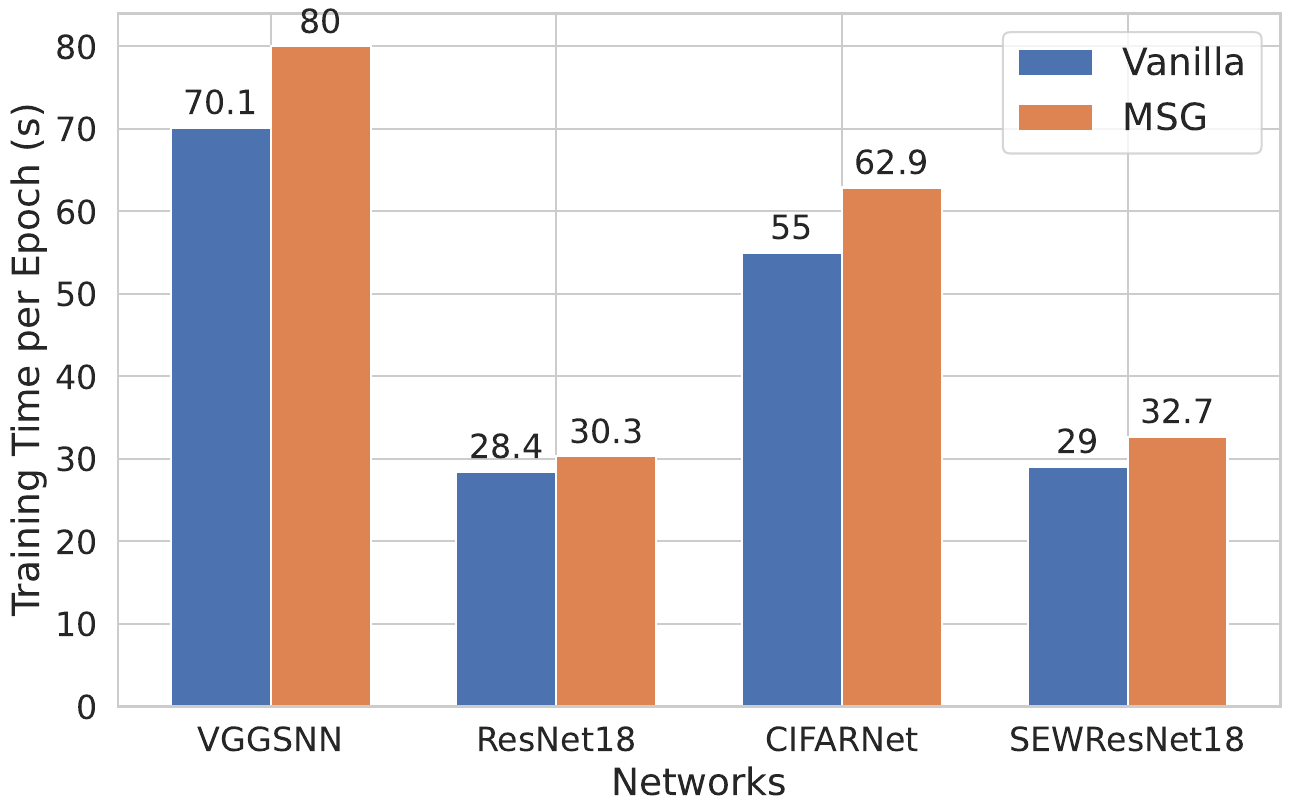}}
    \caption{The firing rate and training time of MSG.}
    \label{sT}
\end{figure*}

\paragraph{The effect of mask probability on MSG} We test the results of MSG on DVS-CIFAR10 with different mask probabilities. Fig. \ref{figp} depicts the comparative results on different network structures (ResNet18 and VGGNet) and surrogate gradients (Arctan and PL Grad). 
The findings suggest that, given varying network architectures and surrogate gradients, an optimal MSG can enhance SNN performance at minimal operational costs. As $p$ nears 1, the network's learning ability markedly diminishes due to the vanishing gradient issue. In contrast, when $p$ tends towards 0, MSG reverts to the conventional surrogate gradient approach, with the gradient mismatch dilemma hindering the SNN's convergence to its ideal solution. Notably, the SNN manifests optimal performance only when $p$ resides within a specific range, commonly between 0.4 and 0.6.

\subsection{Sparsity and Training Time}

In this section, we examine the application of MSG in the training of SNNs. As illustrated in Fig. \ref{sparsity}, the introduction of MSG does not significantly elevate the firing rates across layers, indicating that MSG merely fine-tunes the gradient backpropagation without adding extra spike activity, thereby preserving network sparsity. This is crucial for SNN efficiency as it suggests that MSG enhances training effectiveness and performance without additional computational strain.
Furthermore, Fig. \ref{time} demonstrates that SNN training duration with MSG aligns closely with traditional approaches. This reveals that MSG's optimization of surrogate gradients, facilitated by including a masking operation, does not burden the overall simulation time. Such an approach ensures that MSG maintains the efficacy of surrogate gradients in training SNNs without decelerating the process due to computational complexity.
Overall, integrating MSG into SNN training yields positive outcomes, offering a refined regularization mechanism that preserves network sparsity while ensuring that the new regularization technique does not impede training speed.

\section{Conclusion}

In this paper, we first analyze the update process of the surrogate gradient and propose the Mask Surrogate Gradient method. It combines the SG with the sparsity of original gradient by generating a random mask before each gradient calculation to find a more suitable optimization path for the SNN. In addition, to better decode the output information of the SNN, we propose a temporally weighted output method to enhance the importance of the timestep when single output currents can be accurately classified. Many experiments validate the effectiveness of the proposed algorithm, which outperforms SOTA's algorithm in static and neuromorphic data.

\section*{Acknowledgements}

This study is supported by National Key Research and Development Program (2020AAA0107800).

\section*{Author contributions statement}

Li Yang: Conceptualization, Methodology, Software, Resources, Data Curation, Writing - Original Draft.
Feifei Zhao: Validation, Writing - Review \& Editing, Visualization.
Dongcheng Zhao: Investigation, Writing - Review \& Editing, Visualization.
Zeng Yi: Writing - Review \& Editing, Funding acquisition.

\section*{Competing interests} 

The authors declare no competing interests.

 \bibliographystyle{elsarticle-num} 
 \bibliography{sample}

\end{document}